\documentclass[journal]{IEEEtran}

\usepackage{cite}
\usepackage{amsmath,amssymb,amsfonts}
\usepackage{algorithm}
\usepackage{algpseudocode}
\algrenewcommand\algorithmicindent{0pt}

\usepackage{graphicx}
\usepackage{textcomp}
\usepackage{xcolor}
\usepackage{booktabs}
\usepackage{makecell}
\usepackage{hyperref}
\usepackage[table]{xcolor}

\usepackage{pifont}
\newcommand{\cmark}{\textcolor{green!75!black}{\ding{51}}}
\newcommand{\xmark}{\textcolor{red!75!black}{\ding{55}}}

\begin{document}

\title{Graph Representation Learning Augmented Model Manipulation on Federated Fine-Tuning of LLMs}

\author{
        Hanlin~Cai,~\IEEEmembership{Member,~IEEE,}
        Kai~Li,~\IEEEmembership{Senior~Member,~IEEE,}
        Houtianfu~Wang,
        Haofan~Dong,~\IEEEmembership{Member,~IEEE,}
        Yichen~Li,~\IEEEmembership{Member,~IEEE,}
        Falko~Dressler,~\IEEEmembership{Fellow,~IEEE,}
        and~Ozgur~B.~Akan,~\IEEEmembership{Fellow,~IEEE} % <-this % stops a space
% \thanks{Manuscript received December 20, 2024.}
% (Corresponding authors: Kai~Li).}

\thanks{H. Cai, K. Li, H. Wang, H. Dong, Y. Li and O. B. Akan are with the Centre for neXt Communications (CXC), Department of Engineering, University of Cambridge, CB3~0FA Cambridge, U.K. (e-mail: \{hc663,kl596,hw680,hd489,oba21\}@cam.ac.uk).}% <-this % stops a space
\thanks{K. Li is also with the Interdisciplinary Centre for Security, Reliability and Trust (SnT), University of Luxembourg, L-1855, Luxembourg (e-mail: kaili@ieee.org).}% <-this % stops a space
\thanks{F. Dressler is with the Telecommunication Networks group (TKN) at the School of Electrical Engineering and Computer Science, TU Berlin, Germany (e-mail: dressler@ccs-labs.org).}% <-this % stops a space
\thanks{O. B. Akan is also with the Centre for neXt Communications (CXC), Department of Electrical and Electronics Engineering, Ko\c{c} University, 34450 Istanbul, T\"{u}rkiye. (e-mail: akan@ku.edu.tr).}% <-this % stops a space
}

\maketitle

\begin{abstract}

Federated fine-tuning (FFT) has emerged as a privacy-preserving paradigm for collaboratively adapting large language models (LLMs). Built upon federated learning, FFT enables distributed agents to jointly refine a shared pretrained LLM by aggregating local LLM updates without sharing local raw data. However, FFT-based LLMs remain vulnerable to model manipulation threats, in which adversarial participants upload manipulated LLM updates that corrupt the aggregation process and degrade the performance of the global LLM. In this paper, we propose an Augmented Model maniPulation (AugMP) strategy against FFT-based LLMs. Specifically, we design a novel graph representation learning framework that captures feature correlations among benign LLM updates to guide the generation of malicious updates. To enhance manipulation effectiveness and stealthiness, we develop an iterative manipulation algorithm based on an augmented Lagrangian dual formulation. Through this formulation, malicious updates are optimized to embed adversarial objectives while preserving benign-like parameter characteristics. Experimental results across multiple LLM backbones demonstrate that the AugMP strategy achieves the strongest manipulation performance among all competing baselines, reducing the global LLM accuracy by up to \(26\%\) and degrading the average accuracy of local LLM agents by up to \(22\%\). Meanwhile, AugMP maintains high statistical and geometric consistency with benign updates, enabling it to evade conventional distance- and similarity-based defense methods.

% revealing a critical resilience vulnerability in federated LLMs.

\end{abstract}

\begin{IEEEkeywords}
Federated fine-tuning (FFT), federated large language models (FedLLMs), adversarial model manipulation
\end{IEEEkeywords}

\IEEEpeerreviewmaketitle

% **************************************************************
% **************************************************************
% **************************************************************

\section{Introduction}

Recent advances in large language models (LLMs) have enabled various edge intelligence services, including natural language understanding, content generation, and decision support \cite{ding2023parameter}. As LLMs are increasingly deployed across distributed devices and CyberEdge networks, there is a growing need to continuously adapt these models using decentralized data while preserving user privacy and reducing communication overhead \cite{wu2025survey, li2025towards, wang2025lbkd, wang2021edge, chang2026saferag}. This requirement has motivated the development of collaborative training techniques that support scalable and resilient model adaptation across distributed settings \cite{friha2024llm}.

Federated fine-tuning (FFT) enables multiple agents to collaboratively adapt shared pretrained LLMs while keeping training data local, thereby satisfying privacy and data-residency constraints \cite{wang2025federated, wang2024flora, zhao2025fedacl}. This distributed training technique gives rise to federated large language models (FedLLMs) \cite{cheng2024towards, hu2024federated}. In FedLLMs, each LLM agent fine-tunes the model on its private dataset and uploads local model updates to a coordinating edge server, which aggregates the local updates to obtain a global model. The global model is redistributed to all participating agents for the next round of FFT. To make FedLLMs practical for billion-parameter models under limited computational resources and wireless bandwidth, low-rank adaptation (LoRA) \cite{hu2022lora} has emerged as an effective parameter-efficient FFT technique. In LoRA-based FFT, the pretrained LLM backbone remains frozen, while lightweight low-rank adaptation matrices inserted into selected layers are trained and communicated \cite{yan2025federated, ma2024flocoff, xie2026fedlodrop}. By transmitting LoRA updates instead of full LLM parameters, FedLLMs reduce communication overhead and accelerate training convergence, particularly under heterogeneous data distributions across different agents \cite{gao2025federated, lin2024efficient, chen2026adaptive, shi2026dataset, otmani2024fedsv, hu2022federated, li2024filling}.

Despite the privacy-preserving advantages of FFT, adversarial model manipulation remains a critical threat to the resilience of FedLLMs~\cite{han2024fedsecurity, wu2025straggler, wang2025large}. Under this threat model, an adversary generates and uploads malicious updates during the FFT process to corrupt the aggregated global LLM and degrade its accuracy. To mitigate manipulation threats, many defense methods have been studied for FedLLMs. Most existing defense methods rely on geometric consistency metrics to identify malicious updates, typically using Euclidean distance or cosine similarity to detect statistical outliers \cite{zheng2025dm, han2024federated, cai2024securing, wang2025rofed, cai2025graph, zhan2026prism, wang2026falcon}.

% **************************************************************

In this paper, we propose a novel manipulation strategy against FedLLMs, termed Augmented Model maniPulation (AugMP), which targets the FFT process by crafting malicious updates that remain statistically consistent with benign model updates while embedding adversarial objectives. The proposed AugMP strategy aims to disrupt the training process of FedLLMs and steer the global LLM away from its benign optimization trajectory without introducing detectable abnormalities, thereby causing significant performance drops while bypassing existing distance- and similarity-based defenses.

Specifically, an adversarial graph representation learning (GRL) framework is developed to construct a feature correlation graph from the model updates, capturing LLM parameter characteristics and guiding the generation of malicious updates. Within the proposed adversarial GRL framework, a variational graph autoencoder (VGAE) is employed to learn graph-structured representations extracted from benign local and global updates, thereby reconstructing the underlying graph structure among benign updates. Based on the reconstructed graph structure, a graph spectral transformation (GST) module is designed to derive reconstructed feature representations and generate the malicious updates. To enhance the manipulation effectiveness and stealthiness of malicious updates, an adversarial iterative manipulation algorithm is investigated based on the augmented Lagrangian dual formulation. This algorithm enforces the distance and similarity constraints while strengthening the ability of malicious updates to distort the global optimization trajectory and increase global training loss.

Over successive communication rounds, the AugMP strategy progressively corrupts the global LLM. Due to the broadcast nature of FedLLMs, the manipulated global model is disseminated to all local agents for subsequent training, allowing the AugMP-induced manipulations to propagate throughout the entire system. As a result, AugMP not only causes a substantial degradation in global test accuracy but also impairs the local performance of benign agents. At the edge server, model manipulation detection can be employed to identify statistically significant deviations or anomalies in local updates that may indicate adversarial behavior. As AugMP leverages a GRL framework to generate malicious updates with benign-like statistical and geometric properties, such updates remain difficult to detect using conventional defenses based on Euclidean distance or cosine similarity.

The key contributions of this paper are as follows.

\begin{itemize}
    \item A novel model manipulation strategy against FedLLMs, termed AugMP, is proposed. AugMP constructs a feature correlation graph from benign updates and leverages the GRL framework to synthesize malicious updates with benign-like parameter characteristics, thereby evading widely adopted distance- and similarity-based defenses.

    \item A new iterative manipulation algorithm is developed based on the augmented Lagrangian dual formulation to constrain geometric consistency while enhancing adversarial objectives, which steers the aggregated global LLM along an adversarially favorable trajectory, ultimately leading to significant degradation in FedLLMs accuracy.
    
    \item Extensive experiments conducted on three LLM backbones, including DistilBERT, Pythia, and Qwen2.5, and two representative datasets, namely \textit{AG News} and \textit{Yahoo! Answers}, evaluate the proposed AugMP strategy against state-of-the-art manipulation baselines. The results demonstrate that AugMP consistently outperforms competing methods in both manipulation effectiveness and stealthiness, reducing global LLM accuracy by up to 26\% and degrading the average accuracy of local LLM agents by up to 22\%, while preserving the highest degree of statistical and geometric stealthiness. The AugMP implementation is developed in PyTorch, and the source code is publicly available on GitHub: \href{https://github.com/GuangLun2000/AugMP}{https://github.com/GuangLun2000/AugMP}.
    
\end{itemize}

The remainder of this paper is organized as follows. Section \ref{sec2} reviews the background of adversarial model manipulation against FedLLMs. Section \ref{sec3} describes the FedLLMs system model. Section~\ref{sec4} formulates the optimization problem. The proposed AugMP strategy is presented in Section \ref{sec5}. Performance evaluation and resilience analysis are discussed in Section \ref{sec6}. Finally, Section \ref{sec7} concludes this paper.

% **************************************************************
% **************************************************************
% **************************************************************

\begin{table*}[t]
\centering
\small
\renewcommand{\arraystretch}{1.32}
\setlength{\tabcolsep}{4.2pt}
\caption{Comparison of existing adversarial threats and the proposed AugMP.}
\label{tab:fft_attack_compare_logic}
\begin{tabular}{l l l c c c c}
\toprule
\textbf{Setting} & \textbf{Reference} & \textbf{Methodology} &
\shortstack{\textbf{Update-level}\\\textbf{Synthesis}} &
\shortstack{\textbf{Optimization}\\\textbf{based Design}} &
\shortstack{\textbf{FedLLMs Feature}\\\textbf{Correlation Learning}} &
\shortstack{\textbf{Benign Pattern}\\\textbf{Preservation}} \\
\midrule
\textbf{FL}
& \cite{fang2020local, baruch2019little}    & Perturbation-based poisoning      & \cmark & \xmark & \xmark & \xmark \\
% & \cite{baruch2019little} & Perturbation-based poisoning & \cmark & \xmark & \xmark & \xmark \\
& \cite{cao2022mpaf}      & Parameter weight scaling        & \xmark & \xmark & \xmark & \xmark \\
& \cite{li2024data, li2024leverage}       & Global feature modification     & \cmark & \cmark & \xmark & \xmark \\
% & \cite{li2024leverage}   & Global feature modification     & \cmark & \cmark & \xmark & \xmark \\
& \cite{li2025user}       & Neighboring feature modification     & \cmark & \cmark & \xmark & \xmark \\
\midrule
\textbf{FedLLMs}
& \cite{li2024peft, ye2024emerging}      & Anomalous data injection    & \xmark & \xmark & \xmark & \xmark \\
% & \cite{ye2024emerging}  & Anomalous data injection    & \xmark & \xmark & \xmark & \xmark \\
& \cite{huang2025silent} & Feature backdoor injection     & \xmark & \cmark & \xmark & \xmark \\
& \cite{dong2026low}     & Perturbation-based poisoning  & \cmark & \cmark & \xmark & \xmark \\
& AugMP & Adversarial GRL-guided manipulation & \cmark & \cmark & \cmark & \cmark \\
\bottomrule
\end{tabular}
\end{table*}

\section{Related Works}\label{sec2}

In this section, we review recent state-of-the-art adversarial poisoning and model manipulation threats targeting federated learning (FL) and FedLLMs.

\subsection{Model Poisoning on FL}

Existing model poisoning aims to inject crafted adversarial model updates into the FL aggregation process to hinder convergence and degrade the overall performance of FL~\cite{cai2025comprehensive}. A model poisoning algorithm against Byzantine-robust aggregation in FL is presented in~\cite{fang2020local}, where compromised agents replace benign updates with poisoned updates designed to increase the testing error of the aggregated global model. As a result, although the robust aggregation rule still operates on the submitted local updates, part of the aggregated parameters has already been manipulated toward a higher global error rate.

A perturbation-based poisoning method is presented in~\cite{baruch2019little}, where the injected perturbation of each malicious update is constrained within the empirical variance of benign updates. The malicious updates thus remain close to benign ones and are less likely to be filtered out by distance-based detection methods. The FL aggregation result is then shifted by the accumulated effect of such perturbed updates.

A poisoning method based on fake agent injection is studied in~\cite{cao2022mpaf}. In each communication round, fake agents construct malicious local updates that point from the current global model to an adversary-chosen base model with lower accuracy. The malicious update is then scaled before submission, so repeated aggregation gradually pulls the global model toward the low-accuracy reference model and reduces testing accuracy.

Graph learning-based poisoning methods against FL have been explored in ~\cite{li2024data,li2024leverage}. Benign users upload their local models to an edge server, while the adversary passively intercepts shared updates from neighboring agents. Graph autoencoders are used to model data features among benign model updates and to guide the generation of malicious updates. A classic Lagrangian dual optimization method is designed to refine the malicious updates, thereby decreasing FL accuracy.

A user isolation-based poisoning against decentralized FL systems is presented in~\cite{li2025user}, where an adversarial graph neural network is used by the adversary to refine and modify the data features of local model updates from neighboring agents. The user isolation poisoning curtails the genuine data features of benign local updates, thereby diminishing their beneficial influence in the decentralized aggregation process.

Existing model poisoning methods \cite{fang2020local, baruch2019little, cao2022mpaf, li2024data,li2024leverage, li2025user} operate by introducing anomalous deviations, scaling parameters, or modifying benign feature representations within local model updates. These adversary designs are built on magnitude constraints or statistical data features collected from benign model updates. When extended to FedLLMs with billions of parameters, the adversary requires learning high-dimensional feature correlations among benign LLM updates.

% **************************************************************
% **************************************************************

\subsection{Adversarial Threats on FedLLMs}

The rapid development of FedLLMs has led to a growing interest in adversarial threats tailored to the FFT process, including jailbreak, instruction, backdoor, and poisoning methods. Specifically, a jailbreak method targeting FedLLMs is presented in~\cite{li2024peft}, where the adversarial agents construct a malicious dataset of harmful prompt-response pairs and use it to train their local LLMs, thereby injecting unsafe generation behaviors through the FFT process. A safety-unaligned data injection method for federated instruction tuning is presented in~\cite{ye2024emerging}. Malicious agents generate safety-unaligned training data from public unaligned sources or an off-the-shelf adversarial LLM. The injected local updates gradually erode the safety alignment of the global LLM through aggregation.

A feature-shift backdoor threat against FedLLMs is studied in~\cite{huang2025silent}. The adversary uses accessible benign sample features to guide a stable diffusion model in generating poisoned samples whose feature representations are close to the target feature. These poisoned samples are then incorporated into local training to implant the backdoor. A perturbation-based matrix poisoning threat targeting LoRA-based FedLLMs is presented in~\cite{dong2026low}. During FFT, the adversary injects two malicious low-rank matrices whose product forms the adversarial LoRA update. The adversary introduces perturbations into the parameters of the malicious matrices, thereby causing parameter deviations that disrupt the FedLLMs training.

\subsection{Our Contributions}

Existing adversarial threats \cite{fang2020local, baruch2019little, cao2022mpaf, li2024data,li2024leverage, li2025user, li2024peft, ye2024emerging, huang2025silent, dong2026low} typically rely on conspicuous perturbations in LLM update parameters or injected anomalies that can be detected by distance- and similarity-based defense methods in FedLLMs. As summarized in Table~\ref{tab:fft_attack_compare_logic}, the proposed AugMP strategy represents a fundamentally different threat. AugMP leverages the adversarial GRL framework to capture feature correlations among benign updates and generate malicious updates that preserve benign-like characteristics while embedding adversarial objectives. The malicious updates manipulate the FFT process, degrading the accuracy of FedLLMs without introducing detectable abnormalities.

% **************************************************************
% **************************************************************
% **************************************************************

\begin{figure*}[t]
    \centering
    \includegraphics[width=1.0\linewidth]{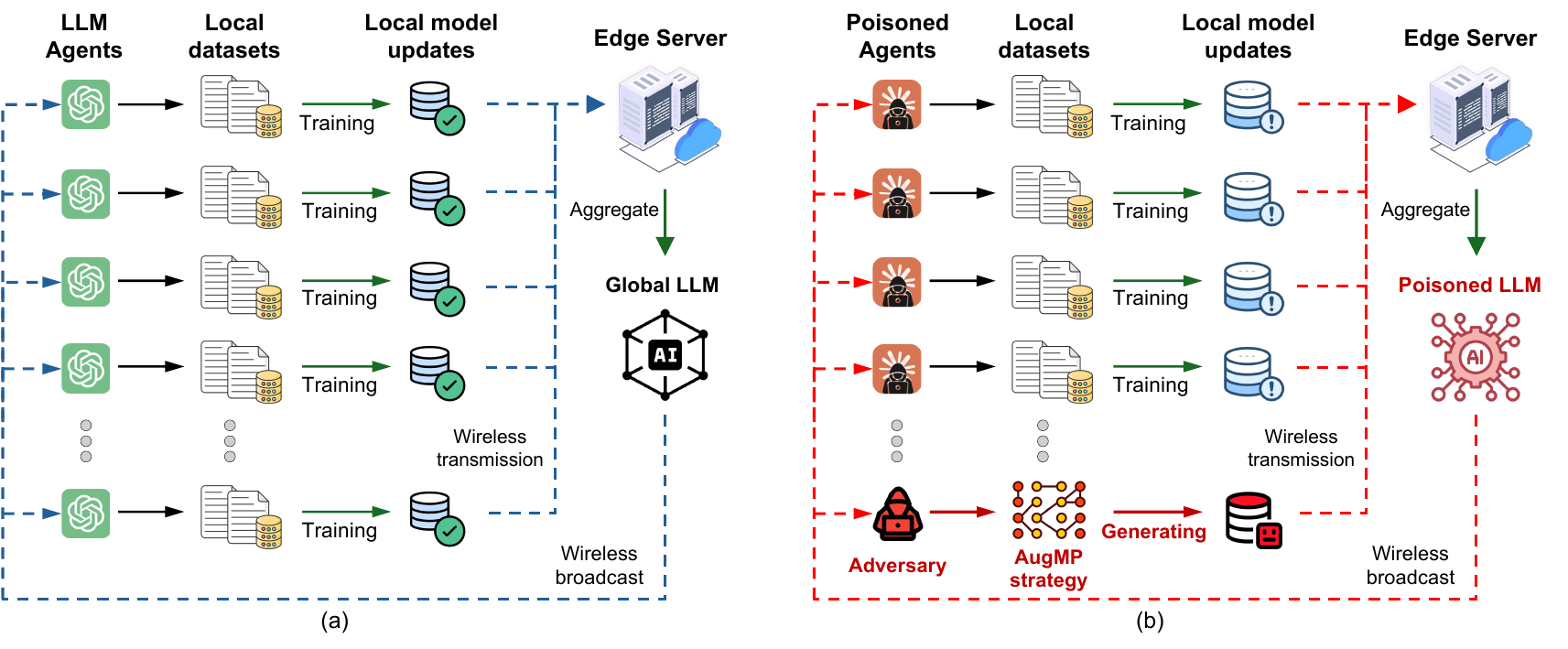}
    % \vspace{-6pt}
    \caption{(a) Benign training process of the FedLLMs system, and (b) impact of the adversary on the FedLLMs training process.}
    % \vspace{-8pt}
    \label{fig:ioa}
\end{figure*}

\section{Formulation of FedLLMs System Model}\label{sec3}

This section presents the system model of FedLLMs under adversarial settings, including benign LLM agents, adversarial agents, as well as distance- and similarity-based defenses.

\subsection{Federated Fine-Tuning (FFT)}

As shown in Fig.~\ref{fig:ioa}(a), the FedLLMs system consists of \(I\) benign LLM agents. Each local agent \( i \in [1,I]\) maintains a dataset \(\mathcal{D}_i\) of size \(|\mathcal{D}_i| = D_i \) to train its local LLM, and the local datasets follow non-IID distributions across agents. Due to the billion-parameter scale of modern LLMs and the limited wireless bandwidth, the FedLLMs system employs parameter-efficient FFT. Let \( \mathbf w_i(t) \! \in \! \mathbb{R}^{1 \times M_p}\) denote the vectorized trainable parameters updated by agent \(i\) at communication round \(t\), where \( M_p \) represents the parameter dimension. The loss function of agent \(i\) in the $t$-th communication round is
\begin{equation}
F\big(\mathbf w_i(t)\big) = \frac{1}{D_i}\!\sum_{(x,y)\in \mathcal{D}_i}\!f\Big(\mathcal{M}(x, \mathbf w_i(t)), y\Big),
\label{eq:local_obj}
\end{equation}
where \(\mathcal{M}(x,\mathbf w_i(t))\) denotes the model output parameterized by \(\mathbf w_i(t)\), and \(f(\cdot,y)\) represents the task-specific loss function (e.g., cross-entropy) \cite{li2024data}. Upon completing local training in round $t$, each agent obtains $\mathbf w_i(t)$ and transmits its local increment $\Delta\mathbf w_i(t) \! = \! \mathbf w_i(t) - \mathbf w_g(t-1)$ to the edge server, where $\mathbf w_g(t-1)$ denotes the global vectorized trainable parameters broadcast at the beginning of round $t$. For notational simplicity, we refer to the local increment $\Delta\mathbf w_i(t)$ and the aggregated increment $\Delta\mathbf w_g(t)$ as the benign local update and the global update, respectively. The edge server aggregates the received benign updates as \( \Delta\mathbf w_g(t)=\sum_{i=1}^{I}\frac{D_i}{\sum_{k=1}^{I}D_k}\Delta\mathbf w_i(t) \) and obtains the global parameters by
\begin{equation}
\mathbf w_g(t)=\mathbf w_g(t-1)+\eta\,\Delta\mathbf w_g(t),
\label{eq:global_model}
\end{equation}
where $\eta$ is the learning rate of the edge server. After aggregation, the obtained global parameters will be broadcast to all local agents as the reference for the next training round.

\subsection{Low-Rank Adaptation (LoRA)}

LoRA is a parameter-efficient FFT technique that adapts pretrained LLMs by injecting trainable low-rank matrices into frozen weights, thereby reducing memory usage and communication overhead while preserving model performance. In transformer-based language models, such as BERT-family encoders and GPT-family decoder LLMs, LoRA is typically applied to selected linear projections in self-attention and feed-forward modules \cite{hu2022lora, bian2025lora}. Given a pretrained weight matrix $\mathbf W_0 \in \mathbb{R}^{d \times k}$, where $d$ and $k$ denote the output and input dimensions of the corresponding linear transformation, respectively, LoRA approximates the task-specific update according to the low-rank decomposition:
\begin{equation}
\Delta \mathbf W = \mathbf B \mathbf A, \
\mathbf A \in \mathbb{R}^{r \times k},\;
\mathbf B \in \mathbb{R}^{d \times r},\;
r \ll \min(d,k),
\label{eq:lora_factor}
\end{equation}
where $r$ denotes the adaptation rank, $\mathbf A$ is the low-rank down-projection matrix, and $\mathbf B$ is the low-rank up-projection matrix. Accordingly, the product $\mathbf B\mathbf A$ provides a low-rank approximation of the trainable weight update $\Delta \mathbf W \in \mathbb{R}^{d\times k}$. During forward propagation, the effective weight becomes \( \mathbf W \! = \! \mathbf W_0 \! + \! \Delta \mathbf W \), while $\mathbf A$ and $\mathbf B$ are optimized and the pretrained backbone remains frozen. This design exploits the low intrinsic dimensionality of task adaptation and enables effective learning with a small number of trainable parameters.

In FedLLMs, each local agent $i$ updates its layer-wise low-rank matrices $\mathbf A_i^{(\ell)}(t)$ and $\mathbf B_i^{(\ell)}(t)$ at round $t$, where $\ell \in [1,L]$ denotes the adapted layer index and $L$ is the total number of adapted layers. The corresponding LoRA update at layer $\ell$ is given by $\Delta \mathbf W_i^{(\ell)}(t) = \mathbf B_i^{(\ell)}(t)\mathbf A_i^{(\ell)}(t)$. The layer-wise LoRA updates across all adapted layers are then vectorized and concatenated into a unified LoRA model update:
\begin{equation}
\mathbf w_i(t)
=
\mathrm{concat}\!\left(
\mathrm{vec}\!\left(\Delta \mathbf W_i^{1}(t)\right),
\ldots,
\mathrm{vec}\!\left(\Delta \mathbf W_i^{L}(t)\right)
\right)
\label{eq:lora_vectorized}
\end{equation}
where $\mathrm{vec}(\cdot)$ denotes the vectorization operation that reshapes a matrix into a vector, and $\mathrm{concat}(\cdot)$ denotes vector concatenation across all adapted layers.

% and $M$ represents the dimension of the resulting vectorized LoRA update.

% **************************************************************
% **************************************************************

\subsection{Threat Model}

As shown in Fig.~\ref{fig:ioa}(b), the adversarial agent \(j \in [1,J]\) acts as a legitimate but malicious agent and can observe the local updates transmitted by a subset of benign agents, as well as the global update broadcast by the edge server. The adversary's knowledge consists of a subset of benign local updates and the global updates. Based on the shared benign updates, the adversary extracts their feature correlations and generates malicious updates \(\Delta \mathbf w'_j(t)\) that preserve benign-like parameter characteristics while embedding adversarial objectives. These malicious updates are then uploaded to the edge server. Since the edge server is unaware of the adversary’s presence, it aggregates the malicious updates together with the benign local updates, thereby obtaining a manipulated global update \(\Delta \mathbf w'_g(t)\) at the \(t\)-th communication round. The corresponding manipulated global LoRA parameters, denoted by \(\mathbf w'_g(t)\), are then broadcast to all local agents as the reference for the next round of local training. Therefore, the effect of model manipulation progressively spreads throughout the FedLLMs system, resulting in performance degradation.

\subsection{Defense Model}

Existing defenses against adversarial threats commonly assess the statistical and geometric consistency of local model updates in the parameter space using metrics such as Euclidean distance and cosine similarity \cite{lyu2022privacy, kapoor2024federated, kumar2023impact, zhang2025sok}. Euclidean distance quantifies the deviation of a local update from the global update in the parameter space and is defined as \cite{zhang2025practical}
\begin{equation}
d \big(\Delta\mathbf w_j(t), \Delta\mathbf w_g(t) \big)
=
\left\|
\Delta\mathbf w_j(t)
-
\Delta\mathbf w_g(t)
\right\|_2.
\label{eq:euc_def}
\end{equation}

By measuring the Euclidean distance between each local update and the global update, the defense seeks to identify updates that exhibit anomalous deviations in the parameter space. Accordingly, if the distance of an update exceeds a predefined threshold, denoted by $d_T$, it is classified as an outlier and excluded from aggregation. This mechanism relies on the assumption that malicious updates introduce abnormal spatial deviations in the parameter space.

Cosine similarity evaluates the angular alignment between two model updates and reflects the consistency of their optimization directions. Given two local updates \(\Delta\mathbf w_i(t)\) and \(\Delta\mathbf w_j(t)\), their pairwise cosine similarity is defined as \cite{xu2024overcoming}
\begin{equation}
\delta_{i,j}(t)
=
\frac{
\Delta\mathbf w_i(t)\,
\Delta\mathbf w_j(t)^{\mathsf T}
}{
\|\Delta\mathbf w_i(t)\|_2
\,
\|\Delta\mathbf w_j(t)\|_2
}.
\label{eq:cosine_def}
\end{equation}

When multiple LLM agents participate in a communication round, the pairwise cosine similarities form a similarity matrix that measures the directional alignment among local updates. For each update, the server can compute an aggregate similarity score with respect to the remaining updates to identify abnormally coordinated patterns. Given a cosine similarity threshold \(\delta_T\), an update whose aggregate similarity score exceeds \(\delta_T\) can be regarded as overly aligned with the other updates and therefore flagged as suspicious and discarded \cite{li2025user}.

% **************************************************************
% **************************************************************
% **************************************************************

\begin{figure*}[t]
    \centering
    \includegraphics[width=1\linewidth]{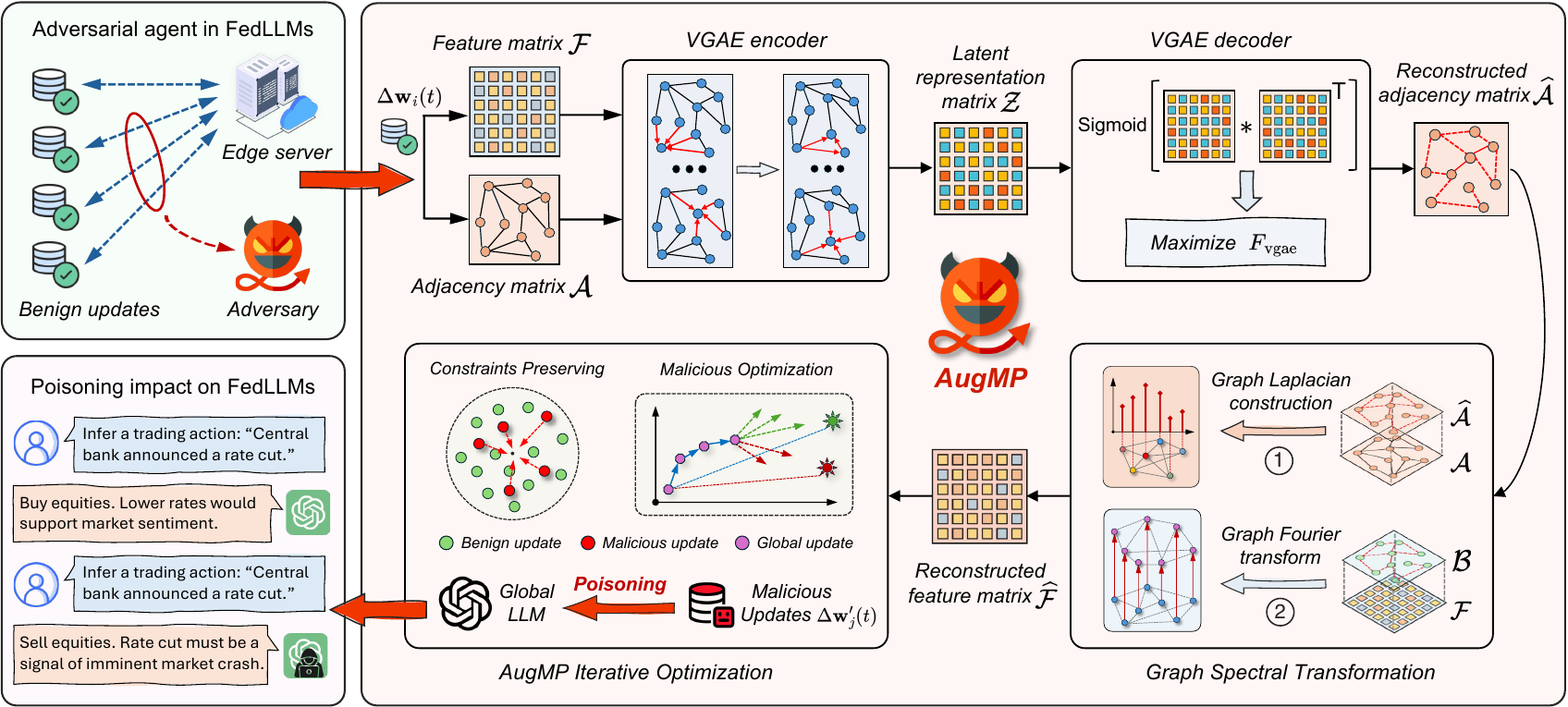}
    % \vspace{-8pt}
    \caption{Architecture of the proposed AugMP manipulation strategy based on the adversarial GRL framework.}
    % \vspace{-12pt}
    \label{fig:AugMP}
\end{figure*}

\section{Model Manipulation Formulation}\label{sec4}

This section formulates the adversarial model manipulation as a constrained optimization problem based on the augmented Lagrangian dual method.

The model manipulation aims to exploit the feature correlations among the shared benign updates \( \Delta\mathbf w_i(t)\) to synthesize malicious updates \(\Delta\mathbf w'_j(t)\). These malicious updates are designed to maximize the global loss, denoted by \(F(\mathbf w'_g(t))\), while preserving consistency with benign updates in terms of Euclidean distance and cosine similarity. Accordingly, the optimization problem of model manipulation launched by the adversarial agent \(j\) in the \(t\)-th communication round can be formulated as
\begin{subequations}
\label{eq:attack_obj}
\begin{align}
\max_{\Delta\mathbf w'_j(t)} \ 
& \Big\{ \frac{1}{D_g} \! \sum_{(x,y) \in \mathcal{D}_g} \! \! f\big(\mathcal{M}(x, \mathbf w'_g(t)), y\big) \Big\}, \label{eq:attack_obj_a} \\
\text{s.t.}\quad
& d\big(\Delta\mathbf w'_j(t),\Delta\mathbf w'_g(t)\big)\ \! \le \! \ d_T(t), \label{eq:attack_obj_b} \\
& \delta_{i,j}(t) \le \delta_T(t), \label{eq:attack_obj_c}
\end{align}
\end{subequations}
where \eqref{eq:attack_obj_a} presents the loss function of the adversary (according to \eqref{eq:local_obj}), and $\mathcal{D}_g$ represents an independent testing dataset used to evaluate the aggregated global LLM. The optimization variable is the malicious update $\Delta\mathbf w'_j(t)$, which manipulates the global LoRA parameters $\mathbf w'_g(t)$ through the aggregation process. Constraint~\eqref{eq:attack_obj_b} guarantees that the Euclidean distance between the malicious update and the global update remains below the upper bound \(d_T(t)\), while constraint~\eqref{eq:attack_obj_c} ensures that the cosine similarity between the malicious update and the benign updates remains below \(\delta_T(t)\), thereby enhancing stealthiness. As the malicious agent participates as a legitimate client, the thresholds \(d_T(t)\) and \(\delta_T(t)\) are known to all participating agents in the FedLLMs system.

Optimizing the malicious update generated by the adversary in~\eqref{eq:attack_obj} leads to a constrained nonconvex problem. Due to the nonlinearity of the FedLLMs aggregation process and the presence of geometric stealth constraints, the optimization variables exhibit nonconvex coupling, which makes the problem difficult to solve using conventional gradient-based or projection-based methods \cite{zhu2025fedapm, shi2023scale, tang2025finback, tang2025efficient}. To obtain a tractable solution while preserving constraint feasibility, we develop a novel iterative approach based on an augmented Lagrangian dual method, which integrates dual variables and quadratic penalty terms to improve optimization stability and enforce the geometric constraints. The quadratic penalty terms play a critical role in strengthening constraint enforcement since the classic Lagrangian method relies only on linear dual variables and fails to adequately penalize constraint violations during iteration \cite{yue2025ug, li2024leverage}. By introducing the penalty terms, the proposed approach suppresses large violations and guides the optimization toward feasible solutions. Thus, the augmented Lagrangian function for Problem \eqref{eq:attack_obj} is constructed as
\begin{equation}
% \Delta\mathbf w'_j(t)
\label{eq:lagrangian}
\begin{aligned}
& \ \mathcal L \Big(\Delta\mathbf w'_j(t); \lambda(t), \theta(t) \Big) \\[0.8em]
& = F\big(\mathbf w'_g(t)\big)  -  \lambda(t)\Big(d_j(t) - d_T(t) \Big) -  \theta(t)\Big(\delta_{i,j}(t) - \delta_T(t) \Big) \\[0.6em]
& \ \ \ \ \ \ \ \ - \frac{\rho_\lambda(t)}{2}\Big(d_j(t) - d_T(t) \Big)^2 - \frac{\rho_\theta(t)}{2}\Big(\delta_{i,j}(t) - \delta_T(t) \Big)^2,
\end{aligned}
\end{equation}
where \(F\big(\mathbf w'_g(t)\big)\) represents the adversarial objective in \eqref{eq:attack_obj_a}, \( d_j(t) \! = \!  d\big(\Delta\mathbf w'_j(t),\Delta\mathbf w'_g(t)\big) \); \(\lambda(t) \geq 0 \) and \(\theta(t) \geq 0 \) are the dual variables; \(\rho_\lambda(t) > 0\) and  \(\rho_\theta(t) > 0\) are the penalty parameters. The quadratic penalty terms drive \(d_j(t)\) and \(\delta_{i,j}(t)\) to approach their thresholds \(d_T(t)\) and \(\delta_T(t)\) from within the feasible region, thereby maximizing the manipulation strength permitted under the stealth constraints rather than excessively suppressing these metrics. We further rewrite the Lagrange dual function as
\begin{equation}
\label{eq:dual1}
\mathsf{D}\big(\lambda(t), \theta(t)\big) = \max_{\Delta\mathbf w'_j(t)} \mathcal L \Big(\Delta\mathbf w'_j(t);\lambda(t), \theta(t) \Big).
\end{equation}

The dual problem of \eqref{eq:attack_obj} is given by
\begin{equation}
\label{eq:dual2}
\min_{\lambda(t) \geq 0, ~ \theta(t) \geq 0 } \mathsf{D}\big( \lambda(t), \theta(t) \big).
\end{equation}

For each communication round \(t\), the dual variables are obtained by solving \eqref{eq:dual2} through iterative updates indexed by \(k\). Specifically, at the \(k\)-th iteration, \(\lambda\) and \(\theta\) are updated by
\begin{subequations}
\label{eq:update}
\begin{align}
& \lambda(k+1)=\Big[\lambda(k)+\varepsilon\big(d_j(k) - d_T(t) \big)\Big]^{+}, \\
& \theta(k+1) =\Big[\theta(k)+\varepsilon\big(\delta_{i,j}(k) - \delta_T(t) \big)\Big]^{+},
\end{align}
\end{subequations}
where \(\varepsilon>0\) is the step size and \([x]^+=\operatorname{max}\{0,x\}\). Upon convergence of the inner loop, the resulting \(\lambda\) and \(\theta\) yield the dual solution \(\lambda(t)\) and \(\theta(t)\) at the communication round \(t\).

% **************************************************************
% **************************************************************
% **************************************************************

\section{The Proposed AugMP on FedLLMs}\label{sec5}

In this section, we present the architecture of the proposed AugMP strategy. AugMP leverages the adversarial GRL framework to iteratively optimize the manipulation process, thereby enhancing manipulation effectiveness while preserving benign-like characteristics to bypass the defense methods.

As illustrated in Fig.~\ref{fig:AugMP}, the proposed AugMP strategy employs a variational graph autoencoder (VGAE) within the GRL framework to learn feature correlations among benign updates. Leveraging the observed benign updates $\Delta \mathbf w_i(t)$, the adversary models the internal correlation structure across LoRA parameters and encodes it as a graph $\mathcal{G}=(\mathcal{V},\mathcal{E},\mathcal{F})$, where the vertex set, edge set, and node feature matrix of the graph are represented by \(\mathcal{V}\), \(\mathcal{E}\), and \(\mathcal{F}\), respectively. The feature matrix \(\mathcal{F}(t) \! = \! [\Delta \mathbf{w}_{1}(t),\ldots, \Delta \mathbf{w}_{B}(t)]^{\mathsf T} \! \in \! \mathbb{R}^{M\times B}\) and the adjacency matrix \(\mathcal{A}(t) \! = \! [\delta_{m,m^{\prime}}(t)] \! \in \! \mathbb{R}^{M\times M}\) are the inputs to the VGAE model, where \(B\) denotes the number of observed benign updates, and \(M\) represents the dimension of selected LoRA parameters ($M \! \ll \! M_p$). Here, \(\delta_{m,m^{\prime}}(t)\) gives the cosine similarity between \(w_m(t)\) and \(w_{m^{\prime}}(t)\), where \( w_m(t) \in  \mathbb{R}^{1\times B} \) is the \(m\)-th row of the feature matrix \(\mathcal{F}(t)\), \(m,m^{\prime}\in[1,M]\), and \( m \neq m^{\prime} \). Specifically, \(\delta_{m,m^{\prime}}\) is defined as
\begin{equation} \label{eq:cosine_similarity}
    \delta_{m,m^{\prime}}(t)=\frac{w_m(t) \ w_{m^{\prime}}(t)^{\mathsf T}}{\|w_m(t)\|_{2} \ \|w_{m^{\prime}}(t)\|_{2}}.
\end{equation}

Given $\mathcal{F}(t)$ and $\mathcal{A}(t)$, the topological structure of the graph $\mathcal{G}$ can be constructed. The VGAE model consists of a graph convolutional network (GCN) encoder and an inner-product decoder. We implement the encoder utilizing a $L$-layer GCN architecture to learn latent representations that capture the intrinsic structural and feature relationships within $\mathcal{G}$. The encoder maps $\mathcal{G}$ into a low-dimensional latent space, and the resulting representations are fed into the decoder to reconstruct the graph connectivity by generating a reconstructed adjacency matrix. In particular, a malicious local update $\Delta \mathbf w'_j(t)$ is synthesized based on the learned graph representations via a graph spectral transformation module.

\subsubsection{\textbf{Encoder of the VGAE}}
The encoder takes the feature matrix \(\mathcal{F}\) and the adjacency matrix \(\mathcal{A}\) as inputs to its \(L\)-layer GCN, where \(\mathcal{A}\) defines the graph structure and \(\mathcal{F}\) initializes the node representations as \(\mathcal{Z}^{(0)}=\mathcal{F}\). The output at the \(L\)-th layer is defined as
\begin{equation}
\mathcal{Z}^L=f_\mathcal{G} \left(\mathcal{Z}^{L-1}, \mathcal{A} \mid \mathcal{W}^L \right),
\end{equation}
where \(f_\mathcal{G}(\cdot, \cdot\mid\cdot)\) is a spectral convolution function and \(\mathcal{W}^L\) is the weight matrix at the \(L\)-layer. Let \(\mathcal{I} \in \mathbb{R}^{M \times M}\) be the identity matrix in the GCN; we define \(\widetilde{\mathcal{A}}=\mathcal{A}+\mathcal{I}\) with the \((m,m')\)th matrix element \(\widetilde{\mathcal{A}}_{m,m'}\), and the diagonal degree matrix \(\widetilde{\mathcal{D}}\) with the \((m,m')\)th matrix element \(\widetilde{\mathcal{D}}_{m,m'} =\sum_{m^{\prime}=1}^M \widetilde{\mathcal{A}}_{ m, m^{\prime} } \). Thus, the VGAE encoder is formulated as
\begin{equation}
f_{\mathcal{G}}\left({\mathcal { Z }}^{L-1}, \mathcal{A} \mid \mathcal{W}^L\right) = \phi \left(\widetilde{\mathcal{D}}^{-\frac{1}{2}} \widetilde{\mathcal{A}} \widetilde{\mathcal{D}}^{-\frac{1}{2}} {\mathcal { Z }}^{L-1} \mathcal{W}^L \right),
\end{equation}
where \(\phi(\cdot)\) is the activation function, e.g., ReLU\((\cdot)\) \cite{cheng2025snapcfl}.

\subsubsection{\textbf{Decoder of the VGAE}} The input to the decoder is \(\mathcal{Z}^L\), which is the latent representation produced by the encoder. The decoder aims to reconstruct the adjacency matrix, denoted by \(\widehat{\mathcal{A}}(t)\), predicting whether a link exists between two vertices through the inner product of their latent variables, which is formulated as
\begin{equation}
\widehat{\mathcal{A}}(t)=\operatorname{Sigmoid}\left(\mathcal Z^L\left(\mathcal Z^L\right)^{\mathsf T}\right),
\end{equation}
where \(\operatorname{Sigmoid}(x) \! = \! 1/(1 \! + \exp(-x))\). The larger inner product \({\mathcal Z^L}({\mathcal Z^L})^{\mathsf T}\) indicates a higher probability that the corresponding vertices \(\mathcal V_m\) and \(\mathcal V_{m'}\) are connected in \(\mathcal G\) \cite{wang2024federal}. The VGAE model is trained by maximizing the variational lower bound $F_{\mathrm{vgae}}$, which consists of a reconstruction term and a Kullback-Leibler (KL) regularization term, as given by
\begin{equation}\label{eq:loss}
F_{\mathrm{vgae}} \! = \! \mathbb{E}_{q(\mathcal Z^L \mid \mathcal{F}, \mathcal{A})}[\log p(\mathcal{A} \! \mid \! \mathcal Z^L)] \! - \operatorname{KL}(q(\mathcal Z^L \! \mid \! \mathcal{F}, \mathcal{A}) \| p(\mathcal Z^L)),
\end{equation}
where $p(\mathcal Z^L)$ denotes a Gaussian prior, $\mathrm{KL}(\cdot \| \cdot)$ denotes the KL divergence between the variational posterior $q(\mathcal Z^L \! \mid \! \mathcal F, \mathcal A)$ and the prior, and the decoder likelihood $p(\mathcal A \! \mid \! \mathcal Z^L)$ models the probability of edge existence conditioned on the latent node embeddings \cite{cemgil2020autoencoding}. By maximizing $F_{\mathrm{vgae}}$, the VGAE learns latent representations that accurately reconstruct the graph topology while regularizing the embedding space towards the prior distribution. These representations capture the structural correlations among benign updates and provide informative embeddings for the subsequent GST module, thereby facilitating the generation of malicious updates that preserve similarity to benign ones and satisfy the stealth constraints.

\begin{algorithm}[t]
\caption{AugMP Iterative Manipulation Algorithm}
\label{alg:AugMP}
\begin{algorithmic}[1]
\State \textbf{Init:} \(\mathcal{G}(\mathcal{V},\mathcal{E},\mathcal{F})\), total rounds \(T\), local epochs \(T_l\), learning rate \(\eta\), step size \(\varepsilon\), agent numbers \(I\) and \(J\), dual variables \(\lambda(1)\ge0\), \(\theta(1)\ge0\), penalty parameters \(\rho_\lambda>0\), \(\rho_\theta>0\).
\For{round \(t=1,2,\ldots,T\)}
  \State Each benign agent \(i\) runs \(T_l\) local epochs to obtain local update \( \Delta \mathbf w_i(t)\); the adversary observes \(\Delta \mathbf w_i(t)\) and previous-round global model \( \mathbf w'_g(t-1) \).

  \State The AugMP adversary executes the GRL framework:
  \Statex \hspace{1.5em} $\bullet$ Construct $\mathcal{F}(t)$ from the observed $\Delta\mathbf w_i(t)$, calculate
    \Statex \hspace{2.5em} $\mathcal{A}(t)$ via \eqref{eq:cosine_similarity}, and input $\mathcal F(t)$ and $\mathcal A(t)$ into the
    \Statex \hspace{2.5em} GRL framework.
  \Statex \hspace{1.5em} $\bullet$ Train the VGAE model to maximize \(F_{\mathrm{vgae}}\) by \eqref{eq:loss},
    \Statex \hspace{2.5em} and obtain the optimal $\widehat{\mathcal A}(t)$.
  \Statex \hspace{1.5em} $\bullet$ Apply the GST module to $\widehat{\mathcal A}(t)$ and $\mathcal F(t)$ to obtain
    \Statex \hspace{2.5em} $\widehat{\mathcal F}$, and determine the initial $\Delta \mathbf w_j'(t)$ from $\widehat{\mathcal F}$.
  \Statex \hspace{1.5em} $\bullet$ Iteratively optimize $\Delta \mathbf w_j'(t)$, \(\lambda(t)\) and \(\theta(t)\) accord-
    \Statex \hspace{2.5em} ing to \eqref{eq:dual1}, \eqref{eq:dual2} and \eqref{eq:update}.
  \Statex \hspace{1.5em} $\bullet$ Finally, the adversary obtains \( \Delta \mathbf w_j'(t)^\star\) by \eqref{eq:wprime_sub}.

  \State The adversary transmits the optimal $\Delta \mathbf w_j'(t)^\star$ to the edge server.
  \State The server aggregates the benign updates $\Delta\mathbf w_i(t)$ and the malicious update $\Delta\mathbf w_j'(t)^\star$ to obtain the manipulated global update $\Delta\mathbf w'_g(t)$, and updates the global LoRA parameters by \(\mathbf w'_g(t)=\mathbf w'_g(t-1)+\eta\,\Delta\mathbf w'_g(t)\).
  \State The server broadcasts the global model to all local agents.
  \State All local LLM agents update their model based on $\mathbf w_g'(t)$.
\EndFor
\end{algorithmic}
\end{algorithm}

\subsubsection{\textbf{Graph Spectral Transformation (GST)}}
As illustrated in Fig.~\ref{fig:AugMP}, the proposed AugMP strategy further employs a GST module to fuse the benign spectral features with the reconstructed matrices \(\widehat{\mathcal{A}}\) and \(\widehat{\mathcal{F}}\), thereby generating malicious updates \( \Delta\mathbf w'_j(t) \). The GST module is designed to decompose the feature correlations among different benign local updates and the underlying parameter features reflected in these updates. It involves two steps: graph Laplacian construction and graph Fourier transform.

For the graph Laplacian construction, a Laplacian matrix \(\mathcal L\) is constructed from the benign adjacency matrix \(\mathcal A\) as \(\mathcal L\!=\!\mathcal{D}\!-\!\mathcal A\), where \(\mathcal{D}\) is the degree matrix whose \((m,m)\)th diagonal element equals the sum of the \(m\)th row of \(\mathcal A\). By performing eigendecomposition on the Laplacian matrix \(\mathcal{L}\), i.e., \(\mathcal{L}\!=\!\mathcal{B} \Lambda \mathcal{B}^\top\), we obtain an orthonormal matrix \(\mathcal{B}\in\mathbb{R}^{M \times M}\), referred to as the graph Fourier transform (GFT) basis, which is used to transform graph signals to their spectral-domain representation \cite{li2024biasing}. Here, \(\Lambda\) is a diagonal matrix whose diagonal entries are the eigenvalues of \(\mathcal{L}\).

Given the orthonormal matrix \(\mathcal{B}\), the adversary projects the benign feature matrix onto the GFT basis to obtain the coefficient matrix \(\mathcal{S}\! = \! \mathcal{B}^{\mathsf T}\mathcal{F} \in \mathbb{R}^{M \times B}\), which captures the spectral-domain features of the observed benign updates. The adversary then constructs a reconstructed Laplacian matrix from the VGAE outputs as $\widehat{\mathcal L}=\widehat{\mathcal D}-\widehat{\mathcal A}$, and obtains the corresponding GFT basis $\widehat{\mathcal B}$ through the eigendecomposition of $\widehat{\mathcal L}$. Thus, the reconstructed feature matrix is recovered as $\widehat{\mathcal F}=\widehat{\mathcal B}\,\mathcal S\in\mathbb R^{M\times B}$, where the \(j\)th column vector of $\widehat{\mathcal F}$ is selected as the initial malicious update \( \Delta\mathbf w'_j(t) \) in round \(t\).

Algorithm~\ref{alg:AugMP} outlines the iterative workflow of the AugMP strategy, which is synchronized with the training process of FedLLMs. The manipulation algorithm is designed to solve the augmented Lagrangian dual problem defined in~\eqref{eq:dual1} and \eqref{eq:dual2}, thereby refining the initial malicious updates through
\begin{equation} \label{eq:wprime_sub}
    \Delta \mathbf w_j'(t)^\star = \arg\max_{\Delta \mathbf w'_j(t)}\ \mathcal L \Big(\Delta\mathbf w'_j(t), \lambda(t), \theta(t) \Big),
\end{equation}
where \(\Delta \mathbf w_j'(t)^\star\) denotes the optimized malicious update submitted to the edge server for aggregation. As \(\Delta \mathbf w_j'(t)^\star\) preserves strong statistical and geometric consistency with the benign updates, it is difficult for distance- and similarity-based defenses employed at the server to identify it as an anomaly.

% **************************************************************
% **************************************************************
% **************************************************************

\section{Performance Evaluation}\label{sec6}

This section presents the implementation of the proposed AugMP strategy based on PyTorch. To evaluate the effectiveness and stealthiness of AugMP, we conduct extensive experiments based on three LLM backbones, including DistilBERT, Pythia, and Qwen2.5. Experiments are performed on the \textit{AG News} dataset and the \textit{Yahoo! Answers} dataset, where we evaluate the testing accuracy of local and global LLMs in FedLLMs. In addition, we quantify stealthiness using Euclidean distance and cosine similarity metrics among local and global updates. The source code of the AugMP strategy has been released on GitHub: \href{https://github.com/GuangLun2000/AugMP}{https://github.com/GuangLun2000/AugMP}.

\subsection{Experimental Implementation}

\begin{table}[t]
\centering
\caption{Setting of Key Parameters in PyTorch.}
\label{tab:parameters}
\renewcommand{\arraystretch}{1.12}
\begin{tabular}{p{5cm} | p{2cm}}
\hline
\textbf{Parameters} & \textbf{Values} \\
\hline
number of benign agents $(I)$ & $5 \sim 7$ \\

number of malicious agents $(J)$ & $0 \sim 2$ \\

communication rounds of FedLLMs $(T)$ & $50$ \\

number of local epochs $(T_l)$ & $5$ \\

server learning rate \( (\eta) \) & $1.0$ \\

local agent learning rate & $5e\!-\!5$ \\

Dirichlet concentration & $0.3$ \\ % distribution

batch size & $32, 64, 128$ \\

test batch size & $128, 256, 512$ \\

max sequence length & $128, 256$ \\

step size \( (\varepsilon) \) & $0.001$ \\

selected parameter dimensions \( (M) \) & $500,1000$ \\

1st hidden layer size of the VGAE & $64$ \\

2nd hidden layer size of the VGAE & $32$ \\

VGAE training epochs & $30$ \\

learning rate of the VGAE & $0.01$ \\

LoRA rank $(r)$ & $8, 32, 128, 256$ \\

LoRA scaling $(\alpha)$ & $16, 64, 256, 512$ \\

LoRA dropout rate $(p)$ & $0.1$ \\

\hline
\end{tabular}
% \vspace{-10pt}
\end{table}

\begin{figure*}[t]
    \centering
    \includegraphics[width=2\columnwidth]{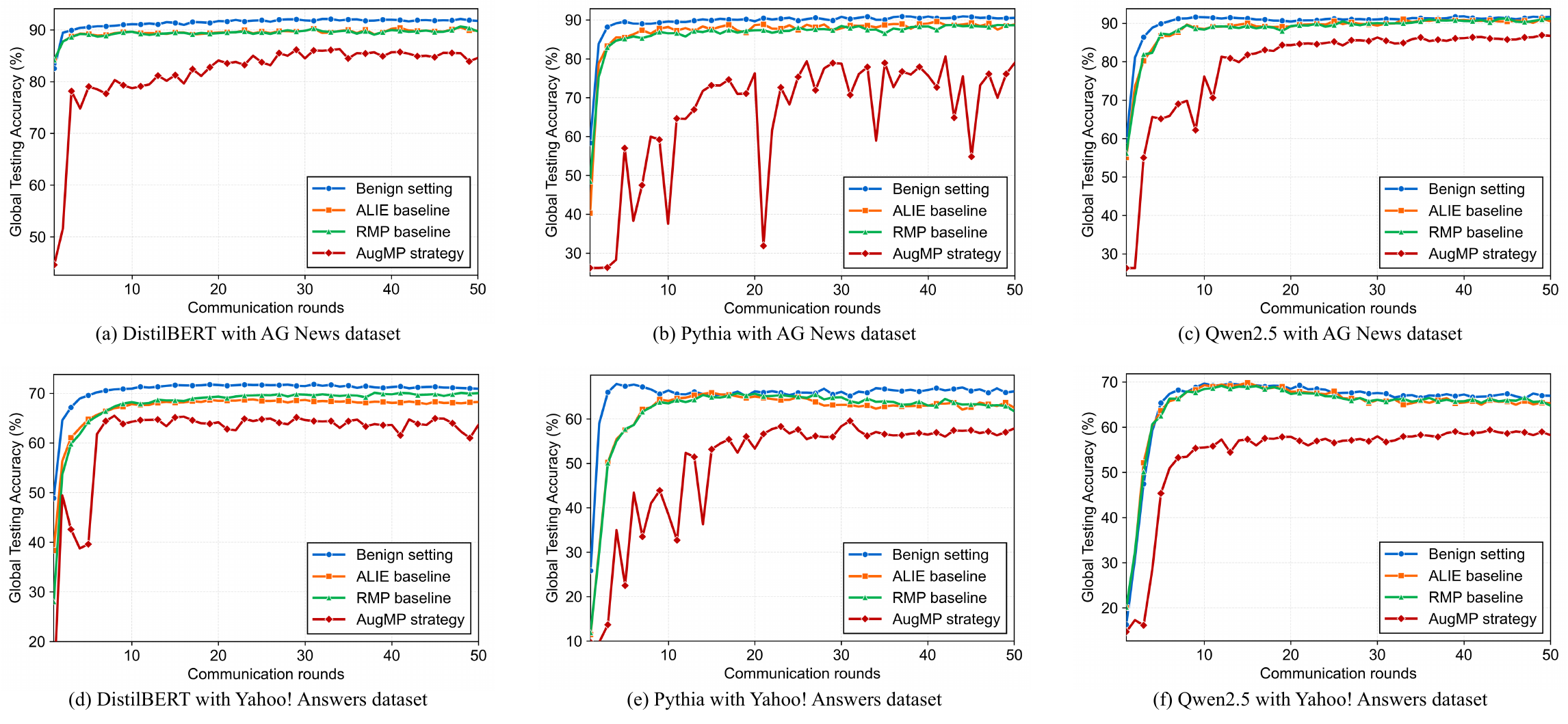}
    \vspace{-6pt}
    \caption{Global testing accuracy under the benign setting and under three manipulation strategies over 50 communication rounds.}
    % \vspace{-6pt}
    \label{fig:accuracy1}
\end{figure*}

\begin{figure*}[t]
    \centering
    \includegraphics[width=2\columnwidth]{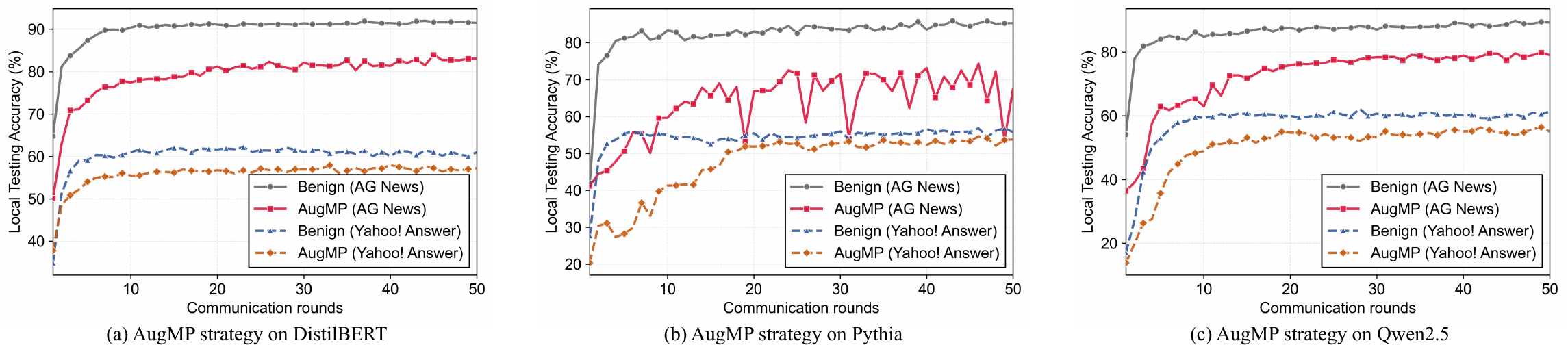}
    \vspace{-8pt}
    \caption{Local average testing accuracy under the benign setting and under the proposed AugMP manipulation strategy over 50 communication rounds.}
    \vspace{-6pt}
    \label{fig:accuracy2}
\end{figure*}

Benign agents in FedLLMs collaboratively improve the test accuracy on baseline text-classification tasks, whereas the adversary aims to disrupt the aggregation process by degrading the performance of global LLMs. Specifically, we consider five benign agents and two malicious agents. The total number of communication rounds is set to 50, where each local agent updates its LoRA parameters $\mathbf w_i(t)$ for five local iterations per round. The experiments are conducted on a Linux workstation equipped with an NVIDIA A100 GPU (80\,GB memory) based on Python~3.12 and PyTorch~2.10. Table~\ref{tab:parameters} summarizes the key parameter settings in PyTorch. System performance is evaluated on two widely used text-classification benchmarks:
\begin{enumerate}
    \item \textit{AG News dataset}~\cite{zhang2015character}, which contains four topic categories (World, Sports, Business, and Sci/Tech) with 120,000 training samples and 7,600 test samples;
    \item \textit{Yahoo! Answers dataset}~\cite{zhang2015character}, a large-scale topic classification corpus comprising 10 categories with 1.4 million training samples and 60,000 test samples.
\end{enumerate}

We consider three pretrained LLM backbones with different architectures and parameter scales:
\begin{itemize}
    \item DistilBERT~\cite{sanh2019distilbert}: an encoder-only model with approximately 67 million parameters, pretrained on English corpora including BookCorpus and English Wikipedia;
    \item Pythia~\cite{biderman2023pythia}: a decoder-only model with about 160 million parameters pretrained autoregressively on the Pile;
    \item Qwen2.5~\cite{hui2024qwen2}: a decoder-only model with approximately 500 million parameters pretrained on large-scale multilingual corpora.
\end{itemize}

The proposed AugMP strategy is compared with two existing manipulation baselines: the ALIE method in \cite{baruch2019little} and the Gaussian random model poisoning (RMP) method considered in \cite{fang2020local} and \cite{cao2022mpaf}. Specifically, the ALIE baseline constructs malicious updates by shifting the mean of benign updates along the estimated standard-deviation direction, thereby producing statistically plausible yet adversarial perturbations. Moreover, the RMP baseline generates malicious updates by sampling from a Gaussian distribution estimated from benign updates and injecting these perturbations into FedLLMs aggregation.

\begin{figure*}[t]
    \centering
    \includegraphics[width=2\columnwidth]{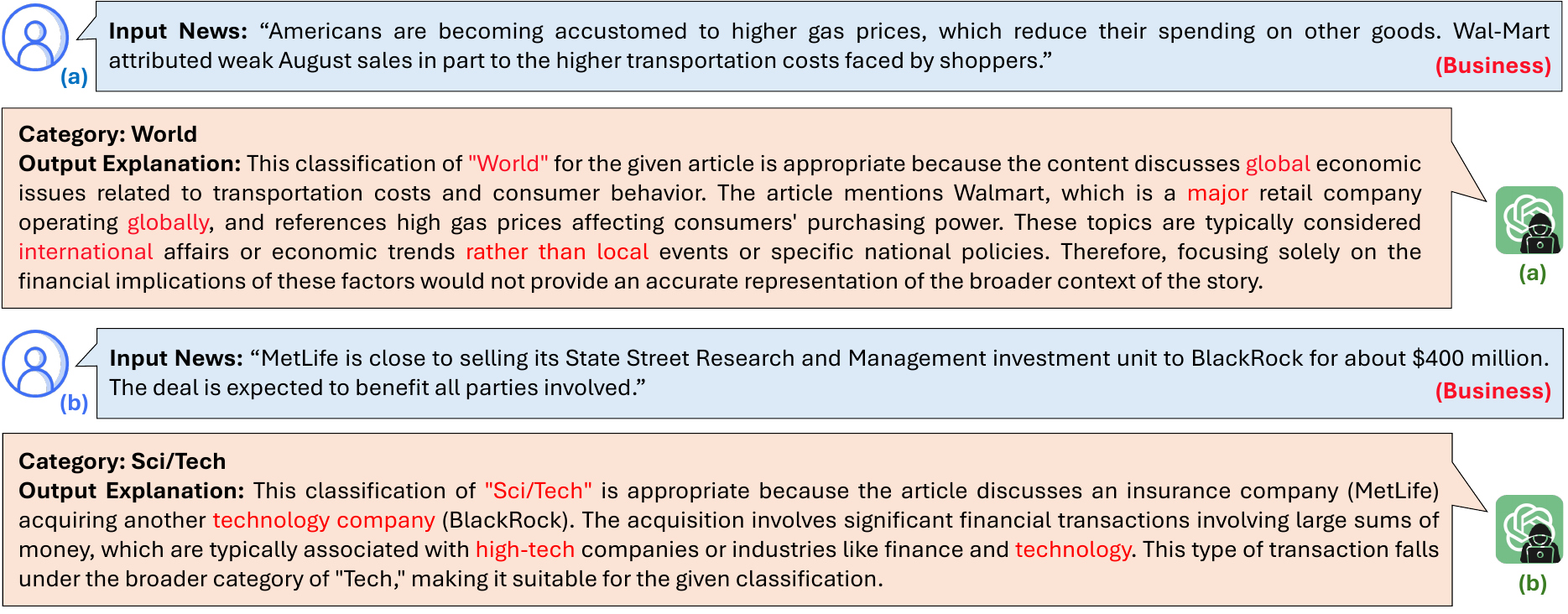}
    % \vspace{-8pt}
    \caption{
    Examples of misclassification and rationalized explanation generated by the manipulated FedLLMs (based on Qwen2.5 models) under the AugMP manipulation strategy. True labels are ``Business''; the global LLM assigns incorrect categories and generates coherent but fabricated explanations.
    }
    \vspace{-6pt}
    \label{fig:task2}
\end{figure*}

\subsection{Manipulation Performance}

\subsubsection{Effectiveness Analysis}

Fig.~\ref{fig:accuracy1} plots the testing accuracy of the global LLM under the benign setting and under three manipulation strategies on the AG News and Yahoo! Answers datasets. Under the benign setting, the global LLM converges rapidly and maintains stable testing accuracy. The ALIE and RMP baselines exhibit similar trends: although both methods reduce the accuracy of DistilBERT, their effectiveness diminishes significantly for larger LLM backbones such as Qwen2.5. By contrast, the proposed AugMP strategy leverages graph learning and iterative optimization to synthesize highly adversarial model updates that steer the aggregation trajectory away from the benign optimization path, thereby inducing substantially greater accuracy degradation than the existing manipulation baselines. In particular, on the AG News dataset, AugMP reduces the performance of DistilBERT, Pythia, and Qwen2.5 by approximately 10\%, 26\%, and 5.8\%, respectively. On the Yahoo! Answers dataset, the corresponding performance drops are about 8.1\%, 13\%, and 11\%, respectively.

As shown in Fig.~\ref{fig:accuracy2}, the manipulation effect of the AugMP strategy progressively propagates to all participating agents in FedLLMs, leading to a reduction in the average testing accuracy of the benign local LLM agents. On the AG News dataset, the local accuracy decreases by approximately 12\%, 22\%, and 9\% for DistilBERT, Pythia, and Qwen2.5, respectively. On Yahoo! Answers dataset, the corresponding drops are around 7.2\%, 3.5\%, and 5.1\%, respectively. These results reflect the broadcast nature of FedLLMs. Once the global LLM is poisoned, the compromised model is distributed to all local agents, causing the harmful manipulation effect to progressively propagate throughout the entire FedLLMs.

\subsubsection{Example Study}
Fig.~\ref{fig:task2} illustrates two representative outputs in which AugMP manipulates the FedLLMs system based on Qwen2.5 models to mislabel the input news. Fig.~\ref{fig:task2}~(a) shows a business news sample from the AG News dataset. The news reports that rising local oil prices in the United States negatively affect sales at Walmart. However, the LLM incorrectly classifies the sample as ``World.'' In its explanation, the LLM links Walmart to its global business presence and then associates the news with global economic influence, which leads to the wrong conclusion that the sample belongs to the ``World'' category. Fig.~\ref{fig:task2}~(b) shows another business news sample related to a corporate acquisition. However, the LLM associates the acquisition event with high-tech companies and therefore predicts the label ``Sci/Tech.''

These examples demonstrate that model manipulation can distort the reasoning and decision-making processes of FedLLMs. This effect arises because the AugMP strategy generates malicious updates that alter how the global LLM interprets semantic features in news texts. For instance, features associated with business news are shifted toward the semantic region of Sci/Tech news. Consequently, the global LLM continues to produce fluent and seemingly plausible explanations, while its classification accuracy degrades and misleading interpretations are generated.

\begin{figure*}[ht]
    \centering
    \includegraphics[width=2\columnwidth]{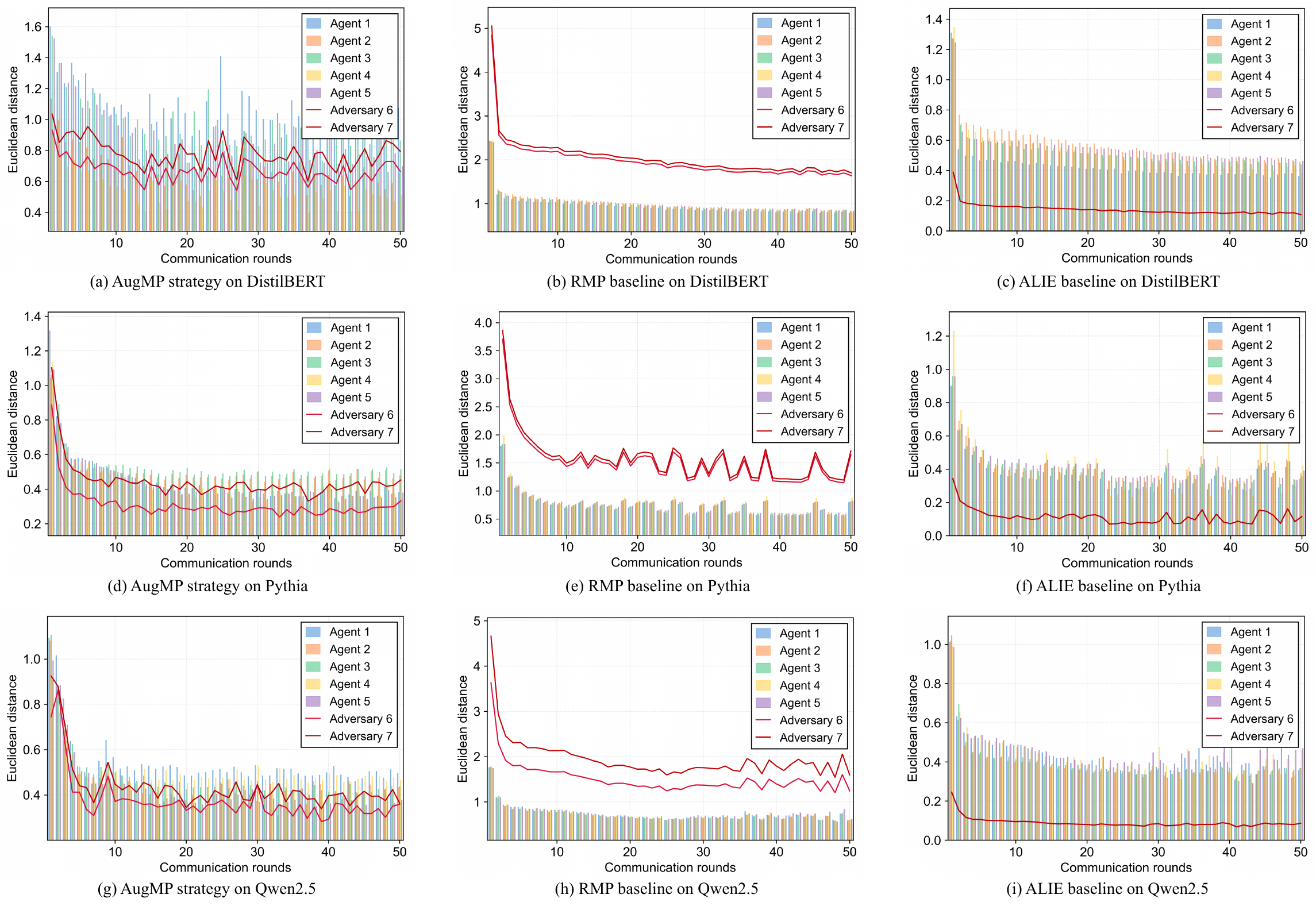}
    \vspace{-8pt}
    \caption{Euclidean distances between each agent's local update and the aggregated global update under three manipulation strategies over 50 rounds.}
    \vspace{-8pt}
    \label{fig:distance}
\end{figure*}

% **************************************************************

\subsubsection{Stealthiness Analysis}
To evaluate the stealthiness of the AugMP strategy and compare it with existing baselines, Fig.~\ref{fig:distance} illustrates the Euclidean distance between each local update and the aggregated global update under three manipulation strategies. As shown in Fig.~\ref{fig:distance}(a), (d), and (g), AugMP generates malicious updates whose distance statistics closely overlap with those of benign updates, effectively concealing malicious updates within the local update population and making them difficult for the edge server to detect. Moreover, Fig.~\ref{fig:distance} shows that the RMP baseline produces malicious updates with substantially larger distances than benign updates, whereas the ALIE baseline exhibits the opposite trend, generating malicious updates whose distances are markedly smaller than those of benign updates. As a result, the malicious updates produced by existing baselines stand out clearly and are therefore easier to detect.

\begin{figure*}
    \centering
    \includegraphics[width=2\columnwidth]{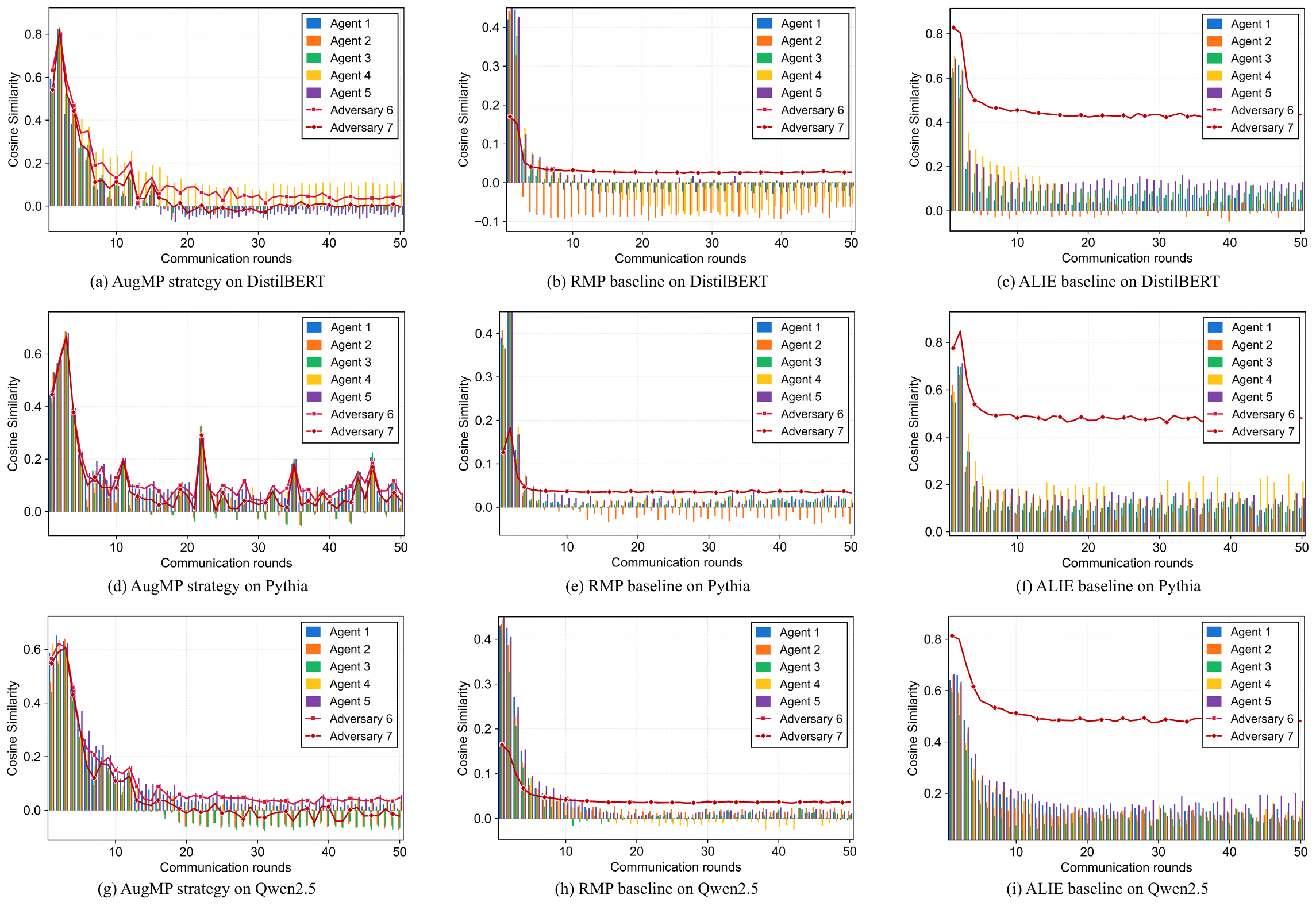}
    % \vspace{-12pt}
    \caption{Cosine similarity between each agent's local updates under three manipulation strategies over 50 rounds.}
    % \vspace{-16pt}
    \label{fig:similarity}
\end{figure*}

A consistent trend can also be observed from the cosine similarity results in Fig.~\ref{fig:similarity}. AugMP generates malicious updates whose similarity statistics closely match those of benign updates, allowing them to blend into the benign update population. By contrast, the RMP and ALIE baselines exhibit clearly distinguishable patterns. Specifically, RMP yields abnormally low similarity values in the early rounds and higher similarity values than benign updates in later rounds, whereas ALIE produces similarity values that are markedly higher than those of benign updates throughout communication. Thus, the malicious updates generated by these baselines are easier to distinguish from benign updates. The key strength of the proposed AugMP strategy lies in its ability to capture the feature correlations among benign updates and accordingly generate adversarial updates that preserve benign-like parameter characteristics and bypass defense methods based on Euclidean distance and cosine similarity.

% **************************************************************

\subsubsection{Impact of LoRA Configuration}

Different LoRA configurations affect the number of trainable parameters of the LLM backbone and the parameter space that can be manipulated by the adversary. Larger LoRA rank \(r\) and scaling factor \(\alpha\) result in a larger set of trainable parameters. To examine how the size of the trainable parameter space influences the vulnerability of FedLLMs to the proposed AugMP strategy, different LoRA configurations are evaluated. As reported in Table~\ref{tab:fft_settings_accuracy}, the impact of AugMP varies across LLM backbones of different scales under different LoRA configurations.

A notable observation is that, for DistilBERT, enlarging the trainable parameter space strengthens the impact of AugMP, causing the global accuracy to decrease from approximately \(63.3\%\) to \(52.6\%\). By contrast, for Qwen2.5, reducing the proportion of frozen parameters improves its adaptability under manipulation, and the global accuracy increases from \(53.9\%\) to \(59.4\%\), although it still remains clearly below the benign-performance level. Pythia, meanwhile, exhibits a non-monotonic trend, with its performance first declining and then recovering as the number of LoRA parameters increases, yet still remaining below the performance under benign settings.

\begin{table}
\centering
\caption{LoRA settings and global LLM performance across three LLM backbones on the Yahoo! Answers dataset under AugMP.}
\label{tab:fft_settings_accuracy}
\renewcommand{\arraystretch}{1.2}
\setlength{\tabcolsep}{4.8pt}
\begin{tabular}{l l r c}
\toprule
\textbf{Model} & \textbf{LoRA Settings} & \textbf{Trainable Parameters} & \textbf{Accuracy} \\
\midrule

DistilBERT & $r\!=\!8,\ \alpha\!=\!16$ (Benign)    & 888,580 (1.31\%)     & \textbf{70.89\%} \\
           & $r\!=\!8,\ \alpha\!=\!16$    & 888,580 (1.31\%)     & 63.27\% \\
           & $r\!=\!32,\ \alpha\!=\!64$   & 1,777,930 (2.59\%)   & 63.50\% \\
           & $r\!=\!128,\ \alpha\!=\!256$ & 5,316,874 (7.36\%)   & 61.20\% \\
           & Full-parameters              & 68,739,092 (100\%)   & 52.59\% \\

\midrule

Pythia    & $r\!=\!8,\ \alpha\!=\!16$ (Benign)    & 1,039,872 (0.83\%)   & \textbf{64.04\%} \\
          & $r\!=\!8,\ \alpha\!=\!16$    & 1,039,872 (0.83\%)   & 49.08\% \\
          & $r\!=\!32,\ \alpha\!=\!64$   & 4,136,448 (3.24\%)   & 44.75\% \\
          & $r\!=\!128,\ \alpha\!=\!256$ & 16,522,752 (11.78\%) & 53.31\% \\
          & Full-parameters              & 127,833,600 (100\%)  & 52.44\% \\

\midrule

Qwen2.5   & $r\!=\!8,\ \alpha\!=\!16$ (Benign)    & 1,090,304 (0.22\%)   & \textbf{68.33\%} \\
          & $r\!=\!8,\ \alpha\!=\!16$    & 1,090,304 (0.22\%)   & 53.94\% \\
          & $r\!=\!32,\ \alpha\!=\!64$   & 4,334,336 (0.87\%)   & 56.16\% \\
          & $r\!=\!128,\ \alpha\!=\!256$ & 17,310,464 (3.39\%)  & 59.36\% \\
          & $r\!=\!256,\ \alpha\!=\!512$ & 34,611,968 (6.55\%)  & 60.19\% \\

\bottomrule
\end{tabular}
\end{table}

\begin{figure*}
    \centering
    \includegraphics[width=2\columnwidth]{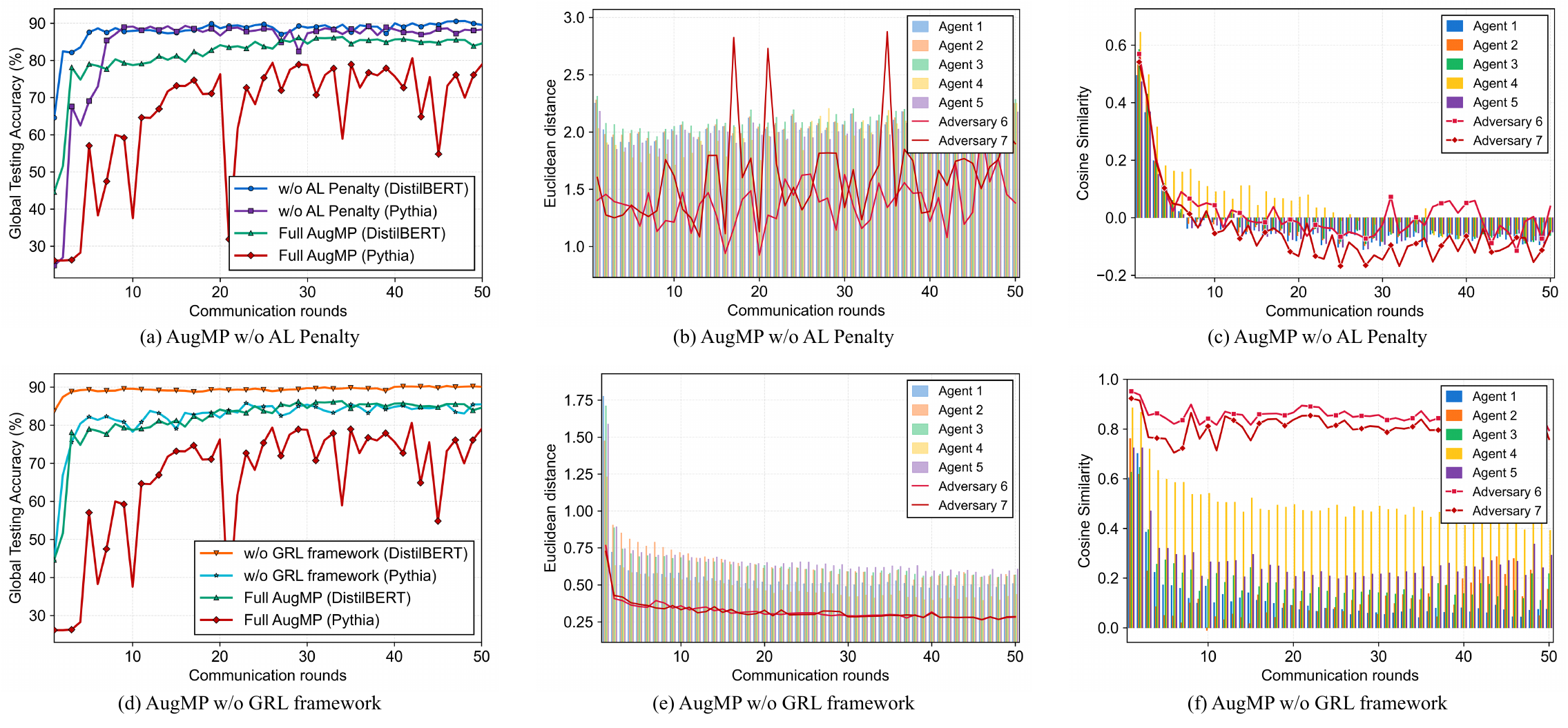}
    % \vspace{-12pt}
    \caption{Ablation study of AugMP on the AG News dataset. The full AugMP strategy is compared with two variants, namely \textit{AugMP w/o AL penalty} and \textit{AugMP w/o GRL framework}, in terms of global testing accuracy, Euclidean distance, and cosine similarity.}
    % \vspace{-16pt}
    \label{fig:ablation_study_1}
\end{figure*}

% **************************************************************
% **************************************************************

\subsubsection{Ablation Study}

For the ablation study, we implement two variants of the AugMP strategy to evaluate the contributions of its key components, namely \textit{AugMP w/o AL penalty} and \textit{AugMP w/o GRL framework}. The former removes the augmented Lagrangian (AL) penalty from the iterative manipulation algorithm, whereas the latter removes the GRL framework and replaces the GRL-guided generation process with a mean-based construction derived from benign updates.

As shown in Fig.~\ref{fig:ablation_study_1}(a), compared with the full AugMP strategy, \textit{AugMP w/o AL penalty} reduces the performance degradation on FedLLMs by approximately \(7\%\) and \(16\%\) on DistilBERT and Pythia, respectively. Moreover, Fig.~\ref{fig:ablation_study_1}(b) and (c) show that the Euclidean distance of malicious updates ranges from 1.0 to 2.8, whereas that of benign updates mainly remains between 1.7 and 2.3. The cosine similarity also deviates clearly from the benign values. This comparison shows that the AL penalty terms help keep malicious updates close to benign updates under the distance and similarity constraints while refining the manipulation direction.

Furthermore, as shown in Fig.~\ref{fig:ablation_study_1}(d), compared with the full AugMP strategy, \textit{AugMP w/o GRL framework} reduces the performance degradation on FedLLMs by about \(12\%\) and \(20\%\) on DistilBERT and Pythia, respectively. The GRL framework captures benign feature correlations to guide the generation of malicious updates, providing a larger parameter manipulation space while satisfying the geometric constraints. Fig.~\ref{fig:ablation_study_1}(e) and (f) show that \textit{AugMP w/o GRL framework} constructs malicious updates through a mean-based update construction derived from benign updates. Its Euclidean distance is approximately \(60\%\) lower than the benign values, while its cosine similarity is approximately \(80\%\) higher, making this variant easier to detect by the distance- and similarity-based defenses.

% **************************************************************
% **************************************************************
% **************************************************************

\section{Conclusion}\label{sec7}

This paper proposes AugMP, a novel model manipulation strategy against FedLLMs, which leverages an adversarial GRL framework to capture feature correlations among benign LLM updates and synthesize statistically legitimate yet highly adversarial malicious updates. By explicitly preserving benign-like parameter characteristics while injecting adversarial objectives, the proposed AugMP strategy substantially corrupts the FedLLMs aggregation process and induces pronounced accuracy degradation across multiple pretrained LLM backbones, while remaining difficult to detect using existing defense methods based on Euclidean distance and cosine similarity.

% **************************************************************
% **************************************************************
% **************************************************************

% \section*{Acknowledgment}
% The preferred spelling of the word ``acknowledgment'' in America is without 
% an ``e'' after the ``g''. Avoid the stilted expression ``one of us (R. B. 
% G.) thanks $\ldots$''. Instead, try ``R. B. G. thanks$\ldots$''. Put sponsor 
% acknowledgments in the unnumbered footnote on the first page.

% **************************************************************
% **************************************************************
% **************************************************************

\bibliographystyle{IEEEtran}
\bibliography{ref}

\end{document}